\documentclass[10pt,twocolumn,letterpaper]{article}

\usepackage[pagenumbers]{wacv} 

\usepackage{graphicx}
\usepackage{amsmath}
\usepackage{amssymb}
\usepackage{booktabs}

\usepackage[pagebackref,breaklinks,colorlinks]{hyperref}

\usepackage[capitalize]{cleveref}
\crefname{section}{Sec.}{Secs.}
\Crefname{section}{Section}{Sections}
\Crefname{table}{Table}{Tables}
\crefname{table}{Tab.}{Tabs.}

\def\wacvPaperID{2695} 
\def\confName{WACV}
\def\confYear{2025}

\begin{document}

\title{SaSi: A Self-augmented and Self-interpreted Deep Learning Approach for Few-shot Cryo-ET Particle Detection}

\author{
  Gokul Adethya \textsuperscript{1}\thanks{Equal contribution} \quad
  Bhanu Pratyush Mantha\textsuperscript{1}\footnotemark[1] \quad
  Tianyang Wang\textsuperscript{2} \quad
  Xingjian Li\textsuperscript{3}\thanks{Corresponding authors} \quad
   Min Xu\textsuperscript{3}\footnotemark[2] \\
  \textsuperscript{1} National Institute of Technology, Tiruchirappalli \\ \quad
  \textsuperscript{2}University of Alabama at Birmingham \\ \quad
  \textsuperscript{3}Carnegie Mellon University
}


\maketitle

\begin{abstract}
Cryo-electron tomography (cryo-ET) has emerged as a powerful technique for imaging macromolecular complexes in their near-native states. However, the localization of 3D particles in cellular environments still presents a significant challenge due to low signal-to-noise ratios and missing wedge artifacts. Deep learning approaches have shown great potential, but they need huge amounts of data, which can be a challenge in cryo-ET scenarios where labeled data is often scarce. In this paper, we propose a novel Self-augmented and Self-interpreted (SaSi) deep learning approach towards few-shot particle detection in 3D cryo-ET images. Our method builds upon self-augmentation techniques to further boost data utilization and introduces a self-interpreted segmentation strategy for alleviating dependency on labeled data, hence improving generalization and robustness. As demonstrated by experiments conducted on both simulated and real-world cryo-ET datasets, the SaSi approach significantly outperforms existing state-of-the-art methods for particle localization. 
This research increases understanding of how to detect particles with very few labels in cryo-ET and thus sets a new benchmark for few-shot learning in structural biology.
\end{abstract}
\bigskip



\section{Introduction}
\label{sec:intro}

Cell biological processes rely on complex networks of molecular assemblies, whose native structures and spatial distributions are crucial to understanding cellular mechanisms. Cryo-electron tomography (cryo-ET) has been gaining popularity in structural biology in recent years. Cryo-ET\cite{irobalieva2016cellular, yahav2011cryo} is a powerful technique that provides three-dimensional (3D) visualization of macromolecular complexes in near-native states at sub-nanometre resolutions. This provides new insights into the understanding of cellular processes and actions of drugs. Cryo-ET has helped in the discovery of many important structures like SARS-Cov-2, which was responsible for the COVID-19 pandemic \cite{liu2020architecture}. Deep learning based automated Cryo-ET image analysis \cite{zeng2020gum,xu2019novo} has received wide attention due to its high efficiency and low cost. However, these computational approaches still face challenges in the localization and classification of 3D particles in cellular environments. First, electrons in imaging interact strongly with biological samples, limiting the dose to prevent damaging samples during imaging. The limited dose limits the resolution of tomograms to about 5nm \cite{koning2010cryo}, which isn't enough to study the structures of macromolecular complexes. Second, imaging angles are typically limited to $\pm$60$^\circ\ $or $\pm$70$^\circ\ $, because of sample thickness. This results in incomplete reconstruction with a missing wedge in Fourier space, making 3D particle picking very challenging compared to 2D particle picking in cryo-electron micrographs. With the detected particles, researchers can further apply subtomogram averaging \cite{briggs2013structural} to enhance the resolution of macromolecular complexes by aligning and averaging copies of identical particles. 

Several existing solutions have been proposed to address the problem of particle localization and classification.
DeepFinder \cite{moebel2021deep} achieved the best localization performance, and Multi-Cascade DS Net \cite{gubins2020shrec} achieved the best classification performance. Recently, DeepETPicker \cite{liu2024deepetpicker} achieved state-of-the-art (SOTA) performance in both classification and localization on the simulated SHREC2021 benchmark. However, given the highly noisy and large 3D tomograms, labeling thousands of particles is too laborious and usually unrealistic in practice. 

This paper considers few-shot learning settings in the problem of particle localization. We find that existing SOTA solutions delivered sub-optimal performance when only tens of particles in a 3D tomogram are manually picked and labeled. This reveals several generalization risks of the existing frameworks, which mostly focus on model architecture and pipeline design. 

To solve this problem, we introduce a Self-augmented and Self-interpreted (SaSi) deep learning approach to achieve few-shot particle detection in 3D cryo-ET images. Based on a general U-Net model architecture, we propose the following novel design, which has not been studied in existing literature. 

\noindent \textit{(1) Self-augmented Volume Infill.} The idea aims to fully exploit the value of the existing available tomogram data to enhance learning. One key challenge is the sparse distribution of particles over the 3D tomogram data, especially for those small structures of interest. This issue is even more severe given only a few particles are labeled. To address this, our first strategy is to increase the particle density. While popular data augmentation techniques only apply a transformation to the existing examples, here we adopt Augmix \cite{hendrycks2019augmix} to simultaneously increase existing particles' variants and volumetric occupancy without relying on external resources. To further make use of the unlabeled tomogram regions, we consider self-supervised learning \cite{pari2021surprising, doersch2015unsupervised, noroozi2018boosting} to learn generalized features. Specifically, we apply contrastive learning. The representations of images are learnt by optimizing contrastive loss utilizing positive and negative pairs, where positive pairs are the pairs of images that are augmentations of the same image, and negative pairs are the augmentations of other images. 

\noindent \textit{(2) Self-interpreted Consistency Guidance.} To further alleviate the dependency on labeled data, we design a novel self-interpreted segmentation approach enforcing the model to interpret and validate its own outputs by leveraging its inherent consistency. For each input example, an augmented version is created using spatial transformations. Instead of fitting the external ground truth, this approach regularizes the segmentation outputs to undergo the same transformation as their corresponding inputs. This component can be used on both supervised and self-supervised learning. This self-validation mechanism helps escape over-fitting to limited ground truth labels for supervised learning. For self-supervised learning complements the standard contrastive learning component by pre-training the decoder of our segmentation model.

Besides the above core techniques, we also improve loss function design and post-processing by considering the particular characteristics of the cryo-ET particle detection task. We use spherical masks instead of ground truth segmentation masks, as they are generally not available in the real world. So, this helps us to generalize better. We evaluate our method's performance on both simulated and real-world datasets.

Our main contributions are summarized as follows:
\begin{itemize}
    \item  We identify the challenge of particle detection in cryo-ET analysis with label scarcity and sparse distribution. To the best of our knowledge, we are the first to explore few-shot learning settings for this challenging task. 
     \item 
     By analyzing the particular characteristics of our target data, we design a novel self-augmented and self-interpreted (SaSi) deep learning approach to solve the problem of few-shot particle detection in cryo-ET images. 
     \item We demonstrate our method's effectiveness on both simulated and real-world cryo-ET benchmark datasets. Compared with state-of-the-art baseline approaches, our method improves the localization performance by a significant margin. 
     \item The core components in our SaSi approach are compatible with most of the popular model architectures, such as CNNs and Vision Transformers (ViT). 
\end{itemize}

\begin{figure*}[t]
  \centering
    \includegraphics[width=0.85\textwidth]{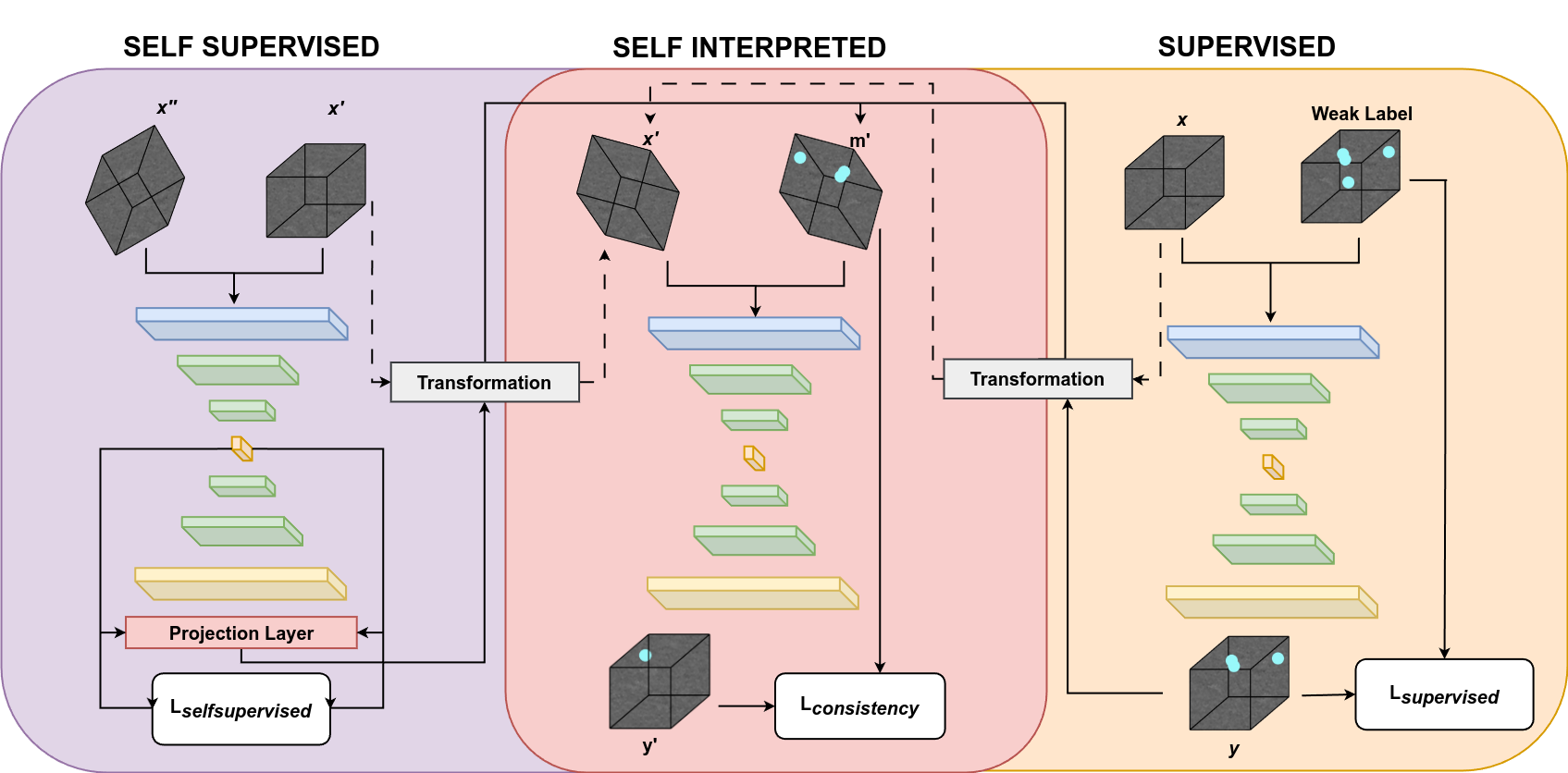} 
    \caption{This figure illustrates the training phase incorporating self-supervised learning using augmented pair \(x', x''\), supervised learning with ground truth mask generated using weak labels, and self-interpreted using \(x'\) and predicted class \(m'\) from either of self-supervised and supervised learning phase.  }
    \label{fig:train}
\end{figure*}

\section{Related Works}
Early works in cryo-ET for localizing macromolecules were based on Template Matching \cite{frangakis2002identification} and DoG (difference of Gaussians) \cite{voss2009dog}.
In Template Matching, the position and orientation of a predefined template are determined by maximizing its cross-correlation with the tomogram., but this has several limitations, like strong dependence on a predefined template and the need for manual threshold tuning.
DoG applies a band passband-pass filter that removes noisy high-frequency components and homogeneous low-frequency
areas, obtaining borders of the structures. The DoG picks particles irrespective of the classes, and the performance depends heavily on the tuning of Gaussian filters.
In recent years, machine-learning methods have been applied to cryo-ET, Chen et al. \cite{chen2012detection} used  Support Vector Machines for detection and classification.

With the increase in cryo-ET data, Deep learning methods started gaining popularity.Che et al. \cite{che2018improved} propose 3 models: DSRF3D-v2, which is a 3D version of VGGNet \cite{simonyan2014very}; RB3D, which is a 3D variant of ResNet \cite{he2016deep} and  CB3D uses a 3D CNN based model for classification of macromolecules. Luengo et al. \cite{luengo2017survos} proposed a supervised approach to classify voxels, but it required manually designed features or rules, which often have various limitations. Chen et al. \cite{chen2017convolutional} developed another supervised segmentation method, utilizing the excellent capabilities of CNN, but a separate CNN is trained for each type of structural feature. Li et al. \cite{li2019automatic} proposed a Faster-RCNN \cite{ren2016faster} based method for automatic identification and localization in a slice-by-slice fashion, but 3D information in adjacent slices was not properly utilized. MC-DS-Net uses a denoising and segmentation architecture; however, they use a real ground truth mask of macromolecules. DeepFinder \cite{moebel2021deep} uses a 3D-UNet for generating segmentation voxels and then applies mean-shift clustering post-processing to find the positions of particles. It also uses spherical masks as weak labels instead of real ground truth masks of macromolecular particles, which are generally unavailable in the real world. DeepETPicker \cite{liu2024deepetpicker} utilizes a 3D ResUNet model, taking advantage of coordinated convolution multiscale image pyramid inputs to enhance localization performance with the help of weak labels. Mean-pooling non-maximum suppression (MP-NMS), post-processing is applied to the generated segmentation output masks to extract the positions of particles. ProtoNet-CE \cite{li2020few} applies a few-shot learning-based method for the task of subtomogram classification only. They combine task-specific embeddings with task-agnostic embeddings, and these combined embeddings are classified using the nearest neighbor classifier. 

\section{Methodology}
In this section, we begin by outlining the problem setup and standard supervised training strategy for the few-shot particle detection task. We also apply self-supervised pre-training to enhance generalization. Then we will introduce the proposed self-augmented volume infill and self-interpreted consistency guidance methods. At last we describe the post processing strategy.  
An overall architecture of SaSi is illustrated in Figure \ref{fig:train}.

\begin{figure*}
  \centering
    \includegraphics[width=0.85\textwidth]{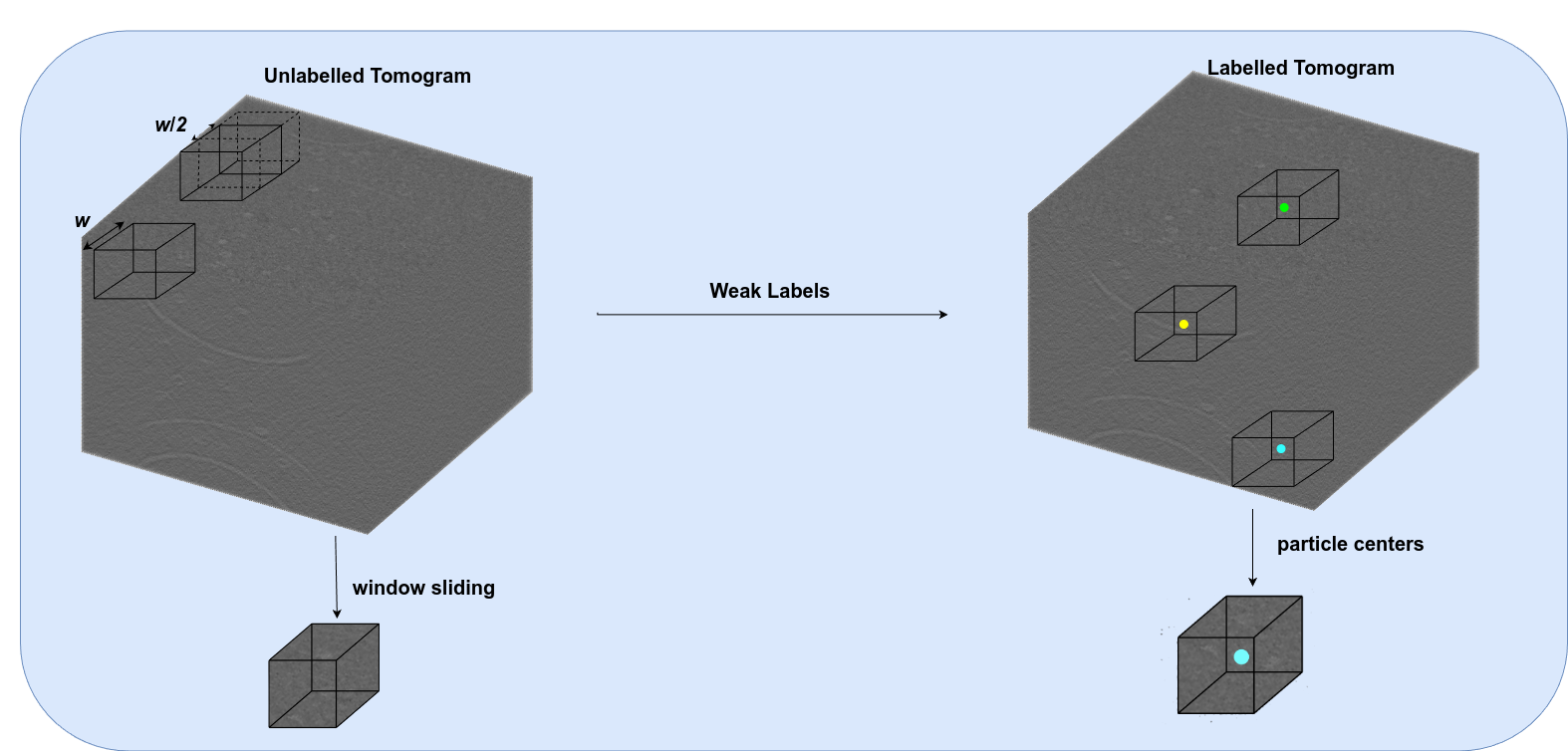} 
    \caption{During testing and self-supervised learning, a sliding window approach with window size W and stride W/2 is applied, while particle center-based sampling is utilized for supervised learning.}
    \label{fig:sampling}
\end{figure*}

\subsection{Problem Setup}

In cryo-ET data analysis, the task of particle detection differs from general object detection in computer vision in several key aspects. Firstly, the original labeling formation for training and evaluation is point annotations to mark particle positions rather than bounding boxes. The objective is to predict the central positions of the particles of interest. This approach is used because the radius is already known for most biological particles. Secondly, only a small number of 3D cryo-ET tomograms are typically collected for a given biological sample. Following standard practice, we use one tomogram for training and the others for testing. It is important to note that, although only a limited number of point annotations are provided for the particles, the entire training tomogram is considered the available dataset since a tomogram represents the smallest unit of data obtained from cryo-ET imaging.

By training a segmentation model with a post-processing step, our final goal is to predict the particle position for an unknown number of particles in the entire tomogram.

\subsection{Supervised Learning} 
To achieve both particle localization and classification, we apply a supervised segmentation model U-Net, which gives voxel-wise predictions. For this part, we mostly follow the practice in existing SOTA approaches \cite{moebel2021deep,liu2024deepetpicker}. 

Since tomogram data have a large spatial dimension, is divided into smaller subvolumes (samples) of size \(W \times W \times W\),  which is the input to the model. To alleviate the potential information loss caused by patch boundaries, we further perform window sliding during inference to obtain subtomograms of size \(W \times W \times W\) using a kernel of size \(W \times W \times W\) and a stride of \(W//2\).

As illustrated in Figure \ref{fig:sampling}, we choose the subtomogram centered on the particle's centroid. Thus, each batch will contain samples of an equal number of particle types. To increase model robustness and prevent loss of information, we use spatial transformation since any other augmentation that affects voxel values can be risky and could distort the underlying pattern. This risk is exceptionally high because the model has fewer examples for counterbalancing the distortion, making it more likely to overfit the altered data.

Using the weak point labels and minimum radius for each particle, we create pseudo-strong labels by generating a sphere of minimum radius filled centered at weak labels centroid. These generated spheres are used as a ground truth segmentation mask, converting the problem into a segmentation task. We adopt Focal Loss and Dice Loss as the learning objective as

\begin{equation}
L_{sup} = \lambda_{dice} L_{dice} + \lambda_{focal}L_{focal},
\label{eq:loss}    
\end{equation}

where the coefficient is set to $\lambda_{dice} = 20$ and $\lambda_{focal} = 1$ according to empirical observations. 

\begin{figure*}
  \centering
    \includegraphics[width=0.85\textwidth]{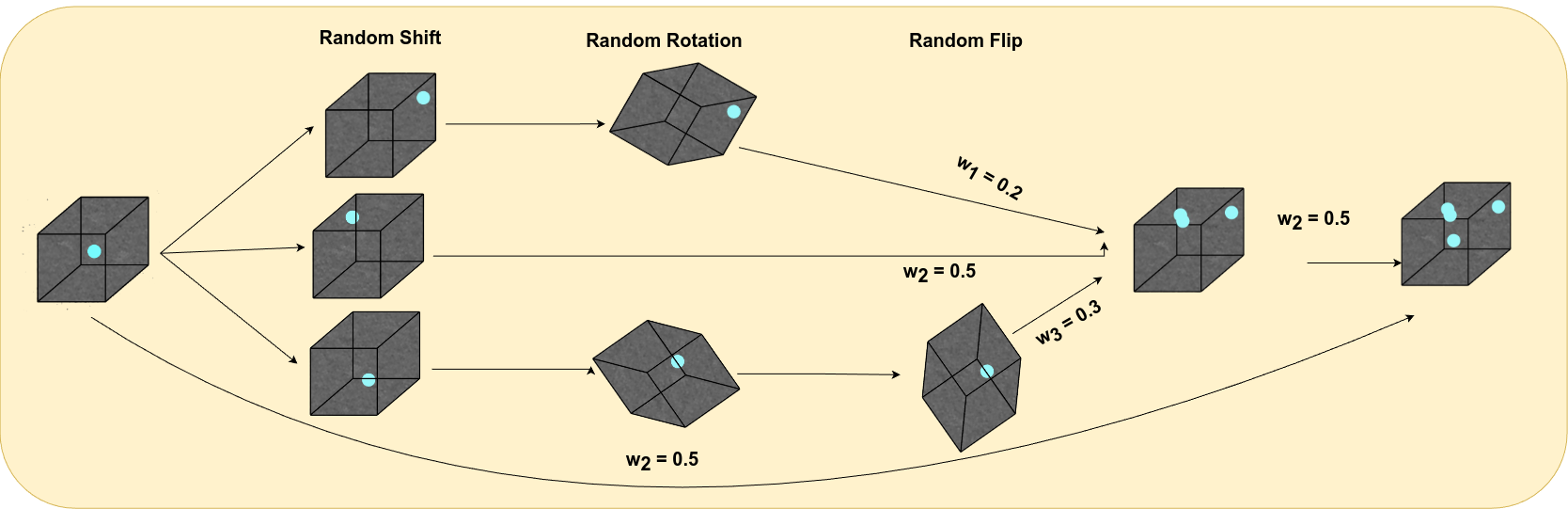} 
    \caption{An illustration of the composition process of our self-augmented volume infill strategy. The input volume is filled with more self-augmented particles with different orientations and positions. }
    \label{fig:augmix}
\end{figure*}

\subsection{Self-Supervised Pre-training}

We use the popular contrastive learning approach SimCLR and NT-Xent Loss on the training subvolumes to learn general cryo-ET image features. 
By passing two augmented variants $x_{mix}^{(1)}$ and $x_{mix}^{(2)}$,  we get their embeddings from the encoder or equivalent block depending on the model architecture and squash them to 1D vectors $e^{(1)}, e^{(2)}$. 
Gupta et al. \cite{gupta2022understanding} has shown the non-linear projection layer to be effective, so we apply a ReLU activation on the squashed embeddings and pass it to a linear layer of output dimension 128 to generate $z^{(1)}, z^{(2)}$. For each input $x$, the NT-Xent Loss maximizes the similarity between its augmented pair of the projected embeddings $z^{(1)}, z^{(2)}$, where minimizes the similarity for different examples.

We perform self-supervised learning during the first few epochs, after which supervised learning continues. The self-supervised learning phase enhances the model's ability to extract robust and meaningful feature representations from the limited available data. After certain epochs, we transition exclusively to supervised learning. This shift allows the model to fine-tune its understanding and adapt specifically to the downstream task, optimizing performance by focusing on task-specific patterns and nuances. This approach ensures that the model benefits from both broad feature learning and targeted task adaptation, leading to improved generalization and efficiency in few-shot scenarios.

\raggedbottom
\subsection{Self-augmented Volume Infill}
Generating augmented examples is crucial in enhancing generalization when training examples are limited. In general computer vision, data augmentation is mainly designed for handling high variability in colors, shapes, and textures across objects. In contrast, cryo-ET images exhibit more challenges about the particle sparsity, where sub-volumes contain few particles buried in high levels of noise. This sparsity complicates particle detection and downstream analysis, especially when datasets are limited. 

We draw inspiration from AugMix \cite{hendrycks2019augmix}, a method originally designed to improve robustness in natural image classification by generating diverse augmentations. However, we adapt AugMix for cryo-ET with a different motivation: instead of addressing variation in visual features, our approach aims to increase particle density within sub-volumes while simultaneously introducing orientation and position variations of the particles. To avoid potential information loss regarding structural changes, we limit the set of augmentation operations $\mathcal{T}=\{T_1, T_2,..., T_n\}$ to include only random shifting, rotation, and flipping. During the AugMix process, an input example $x$ is augmented to several variants $\{x_i^{\prime}\}_{i=1}^m$ by different augmentation chains. Each chain consists of a sequence of $k$ augmentation operations selected randomly from the predefined set $\mathcal{T}$ as $x^\prime_i = T_{i_i} \circ T_{i_2} \circ ... \circ T_{i_k} (x)$. The mixed augmented example is calculated by $x_{aug} = \sum_{i=1}^{m} \mathrm{w}_i x^{\prime}_i $, with each coefficient $\mathrm{w}_i$ randomly sampled from a Dirichlet distribution as $\mathrm{w}_i \sim \text{Dirichlet}(\alpha, \alpha, \ldots, \alpha)$.  We then mix the result of the augmentation chain and the original image at "skip connection" to generate the final example as $x_{mix} = \beta x + (1-\beta)x_{aug}$ with $\beta$ sampled from a Beta distribution. We adopt the same mixing process for the annotation and use the argmax operation to get the final Augmix-generated ground truth mask. This process is illustrated in Figure \ref{fig:augmix}.

\subsection{Self-interpreted Consistency Guidance}
Here we introduce the self-interpreted learning approach for training segmentation models without requiring ground truth segmentation masks. Our method relies on the intuition that a model should be able to explain its predictions instead of only fitting external ground truth labels in a blind manner. We achieve this goal by involving a transformation-aware consistency guidance (CG) loss on predicted segmentation maps. Specifically,  For each input example $x$, we generate an augmented version $x^{\prime}$  by applying a spatial transformation $T$ as $x^{\prime} = T(x)$. The segmentation model $f$ predicts segmentation maps $m = f(x)$ and $m^{\prime} = f(x^{\prime})$ for the original and augmented inputs, respectively. Then we apply the same transformation $T$ on the segmentation map $m$ and use this transformed map $T(m)$ as a target to learn $f(x^{\prime})$. To be specific, we apply the same loss function $L(m^{\prime}, T(m))$ as in Eq. (\ref{eq:loss}).

The rationale behind this self-interpreted approach is that if a model truly understands the structure of the objects it segments, its segmentation results should remain consistent according to the spatial transformations applied to the input. We apply this self-interpreted segmentation loss on both self-supervised learning and supervised learning phases of the training as it requires no labels. In addition, unlike traditional contrastive learning approaches that focus on transformation-invariant feature learning only for the encoder part, our self-interpreted feature learning also covers the decoder.

\subsection{Post Processing} 
We apply the arg max operation on the model output to get a voxel-wise classification. Then we apply connected components in 3D (cc3d) \cite{Silversmith_cc3d_Connected_components_2021} using 26 connected components, which uses a 3D variant of the two-pass method by Rosenfeld and  Pflatz augmented with Union-Find and a decision tree based on the 2D  8-connected work of Wu, Otoo, and Suzuki on this sub tomogram to generate some arbitrary number of clusters. Compared against previous post-processing strategies Meanshift (in Deepfinder \cite{moebel2021deep}) and MP-NMS (in DeepETpicker \cite{liu2024deepetpicker}), we find cc3d is much more reliable in the few-shot settings, and it requires no hyperparameters to be tuned. Therefore, in our experiments, we also integrate cc3d into Deepfinder and DeepETpicker as stronger SOTA baselines.

\begin{table*}[!ht]
\centering
\caption{
{\bf Localization F1 scores on different few-shot settings.} $N$ represents the number of labeled points for each class. SaSi denotes our method built on top of DeepFinder. Note that SaSi uses cc3d as the default post-processing strategy, as it performs better when applied to baseline frameworks.}
\begin{tabular}{|l|c|c|c|c|}
\hline
Method & N=3 & N=5 & N=10 & Avg \\ \hline
DeepFinder (meanshift)&0.088&0.050&0.103&0.080  \\ \hline
DeepFinder (cc3d)&0.221&0.233&0.260&0.238\\ \hline
DeepETpicker (nmsv2)&0.024&0.027&0.023&0.025\\ \hline
DeepETpicker (cc3d)&0.267&0.206&0.276&0.250\\ \hline
SaSi & 0.325 & 0.396 & 0.409 & 0.376\\\hline
\end{tabular}
\label{table1}
\end{table*}

\begin{table*}[htbp] 
\centering
\caption{
{\bf Localization performance evaluated by F1 score on the real world dataset \cite{guo2018situ}}. The test refers to the official test split, and Val+Test refers to using the combination of the test and validation tomogram as the test set.}
\begin{tabular}{|l|c|c|c|c|c|c|}
\hline
Method & \multicolumn{2}{c|}{\(N=3\)} & \multicolumn{2}{c|}{\(N=5\)} & \multicolumn{2}{c|}{\(N=10\)} \\ \cline{2-7} 
& Test & Val+Test & Test & Val+Test  & Test & Val+Test \\ \hline
DF (cc3d) & 0.050	&0.066	&0.045&0.076&0.132&0.123\\ \hline
SaSi (cc3d) & 0.060 & 0.076 & 0.081 & 0.098 & 0.136 & 0.163 \\ \hline
\end{tabular}
\label{tab:real world}
\end{table*}

\section{Experiments}
\subsection{Datasets}
\subsubsection{Simulated Dataset}
We use SHREC2021 \cite{10.2312:3dor.20211307}, which contains ten 3D tomograms, the ground truth containing information about each macromolecule's localization and class. It also contains weak labels in the form of the coordinates of the centroid of each macromolecular particle. Each 3D tomogram is of size $512 \times 512 \times 512$. There are a total of 12 protein classes, vesicles and gold fiducials. 

We represent $i_{th}$ tomogram using $\mathrm{T}_i$. To better simulate the realistic low-resource scenario, we use only the tomogram $\mathrm{T}_0$ as the training data to simulate the few-shot learning scenario. We randomly pick 3, 5, or 10 labeled particles for training, thus creating three few-shot settings. Compared with many previous work that use all the 8 tomograms as training data, our settings align better with realistic scenarios that only a few particles can be annotated by biological experts when analyzing new tomograms.

The SHREC2021 \cite{10.2312:3dor.20211307} dataset officially suggests $\mathrm{T}_9$ as the test data. Considering the high evaluation variance by using only one tomogram as the test data, we adopt two additional sets of tomograms as the test data, which are (1) $\mathrm{T}_{8-9}$ representing tomogram (8,9) and (2) $\mathrm{T}_{6-9}$ representing tomogram (6,7,8,9). For each N-shot scenario, we report the results as the average over $\mathrm{T}_9$, $\mathrm{T}_{8-9}$, and $\mathrm{T}_{6-9}$ for a more reliable evaluation, while still treating the official test data as the most important one, following previous practice. We use the F1 score to evaluate the performance of particle localization.

\subsubsection{Real Dataset}
We use real experimental data \cite{guo2018situ} consisting of 3 tomograms with 3 annotated classes of macromolecular structures, namely RIBO, 26S, and TRIC. There are a total of 646 labeled macromolecules and the size of each tomogram is $375 \times 926 \times 926$. Since we have only 3 tomograms, we use 1 tomogram each for training (G4L3T1), validation (G3L6T1), and test (G4L3T2). Similar to the SHREC dataset, here we use both the official test split and the combination of validation and test set for evaluation. Since the labeled data are extremely scarce in the few-shot scenario, we apply a fixed set of empirical hyperparameters without using the validation set for hyperparameter tuning. 

\subsection{Baselines}

We evaluate DeepFinder's and DeepETPicker's performance using the model architecture as illustrated in their respective papers. For DeepFinder, we tested two different post-processing methods, the first being mean-shift, as given in the paper, and the other being cc3d. Similarly, for DeepETPicker, we use MP-NMS and cc3d post-processing.

\begin{table*}[htbp]
\centering
\caption{Analysis of Volume Infill (VI) on the SHREC benchmark.}
\begin{tabular}{|l|c|c|c|c|}
\hline
Method & N=3 & N=5 & N=10 & Avg \\ \hline
Weight baseline $v(\mathcal{S})$ & 
0.228 & 0.230 & 0.273 & 0.244\\
Weight $v(\mathcal{S} \cup \{\text{VI}\})$ & 0.280 & 0.333 & 0.403 & 0.339\\ 
Marginal contribution $\phi_{\text{VI}}$ & 0.052 & 0.097 & 0.130 & 0.095 \\ \hline
\end{tabular}
\label{tab:results augmix}
\end{table*}

\begin{table*}[htbp]
\centering
\caption{Analysis of Consistency Guidance on the SHREC benchmark.}
\begin{tabular}{|l|c|c|c|c|}
\hline
Method & N=3 & N=5 & N=10 & Avg \\ \hline
Weight baseline $v(\mathcal{S})$ & 0.244	&0.316	& 0.339	& 0.300
\\
Weight $v(\mathcal{S} \cup \{\text{CG}\})$ & 0.265	&0.325	&0.347	&0.313
 \\
 Marginal contribution $\phi_{\text{CG}}$ & 0.021 & 0.009 & 0.008 & 0.013\\\hline
\end{tabular}
\label{tab:consistency_table}
\end{table*}

\subsection{Implementation details}

All our experiments are conducted using PyTorch. We apply random rotation, random flip, and random shift augmentation to increase the data size artificially. In Random Shift, we shift the subtomogram center within the range of $\pm 50\%$ of the subtomogram size. For random rotation, we take a subtomogram of size up to \(\sqrt{2} \times W\) to avoid filling the voxel with arbitrary values and to preserve the voxel information. For Augmix, we set $\alpha=1$ in Dirichlet distributions.
We use the Adam optimizer with a learning rate of 0.0001. The batch size used is 16, and we use a combination of dice loss and focal loss in the ratio 20:1. The subtomogram size is $24\times24\times24$ voxels. So the kernel size used is $24\times24\times24$, and the stride is 12. Considering the few-shot setting, we train each model for a total of 8000 epochs, which is equivalent to around 10,000 to 80,000 iterations for most configurations. 
The temperature parameter used for NT-Xent loss is 0.1. We perform self-supervised learning for 10 epochs, followed by supervised learning until training finishes. The self-interpreted learning loss is applied before the 4000th epoch as it achieved better empirical performance. For training, we use 8xA5000 GPUs. All experimentation for SaSi is conducted using the DeepFinder architecture

\section{Experiments}
\subsection{Main Results}
Table \ref{table1} reports the performance of baselines and our approach on the SHREC 2021 benchmark. It is shown that our SaSi approach outperforms existing SOTA baselines DeepFinder and DeepETPicker on the few-shot settings with significant margins. We observe that the original post-processing strategies suggested in the original papers are not robust in the few-shot learning settings. Using the advanced cc3d strategy brings a significant improvement on DeepFinder and DeepETPicker. We use them as amended baselines. Nevertheless, SaSi even surpasses those stronger SOTA baselines. We also notice that DeepFinder and DeepETPicker shows a certain degree of instability in few-shot settings, e.g., sometimes getting lower F1 score on the N=5 setting than that of N=3. We suspect this may be caused by the high randomness of the few-shot learning problem, i.e., the quality of the few selected labeled particles affect the learning effects a lot. 

From Table \ref{tab:real world}, we also see similar improvements while applying SaSi in real-world datasets.
Though all the results on this challenging real dataset are still low, SaSi shows clear improvements in most experimental groups.

\subsection{Ablation Study}
Since the effects of Self-supervised Pre-training (SSP) have been extensively discussed in existing literature, here we present additional experimental results to analyze the value of our original ideas  Volume Infill (VI) and Consistency Guidance (CG). To evaluate their individual contributions in the SaSi algorithm, we adopt the Shapley value principle \cite{shapley1953value,scott2017unified} from cooperative game theory. The Shapley value provides a fair attribution of the overall performance improvement to each component by considering their contributions both individually and in combination with other components. A positive score means positive contribution to the whole system. 

Specifically, let \( v(\mathcal{S}) \) represent the performance metric (i.e., F1-score in this paper) of the SaSi algorithm on a given dataset, where \( \mathcal{S} \subseteq \{\text{SSP}, \text{VI}, \text{CG}\} \) denotes a subset of the algorithm's components. Our goal is to compute the Shapley values for VI and CG, which measure their marginal contributions to the overall performance. The Shapley value for a component \( i \in \{\text{VI}, \text{CG}\} \) is defined as:

\[
\phi_i = \underset{\mathcal{S} \subseteq \{\text{SSP}, \text{VI}, \text{CG}\} \setminus \{ i \}}{\sum} \frac{|\mathcal{S}|! \cdot (n - |\mathcal{S}| - 1)!}{n!} \left[ v(\mathcal{S} \cup \{i\}) - v(\mathcal{S}) \right].
\]

Here  \( v(\mathcal{S} \cup \{i\}) - v(\mathcal{S}) \) represents the marginal contribution of component \( i \) to subset \( \mathcal{S} \), and  \( |\mathcal{S}|! \cdot (n - |\mathcal{S}| - 1)! / n! \) is the weighting factor ensuring that each subset is fairly considered based on its size. 

The results of Shapley value analysis are presented in Table \ref{tab:results augmix} and Table \ref{tab:consistency_table}. For easier observation, we also include the average F1 scores for weighted baselines $v(\mathcal{S})$ and the corresponding variations with the target component $i$, i.e., $v(\mathcal{S} \cup \{i\})$ . The two tables show consistently positive Shapley values for  $\phi_{\text{VI}}$ and  $\phi_{\text{CG}}$, indicating that both components contribute positively to the few-shot particle detection framework. We also observe that Volume Infill has a larger influence on performance than Consistency Guidance. This is likely because VI addresses the sparsity issue directly by increasing the particle density in cryo-ET sub-volumes, which becomes more impactful as more samples are available for training. On the other hand, CG still provides a steady regularizing effect which ensures that the model learns more consistent representations across augmented data. Both components make complementary contributions in the SaSi algorithm.

Another observation which seems counterintuitive is that, VI shows more improvements under fewer-shot settings. We suspect this can be potentially explained by  the  extremely poor data diversity in fewer-shot settings such as N=3. Specifically, if the initial dataset is too small and homogeneous, there may be insufficient diversity for VI to generate meaningful variations that improve generalization, although the volume density is indeed increased. However, when  the training data has more intrinsic diversity as involving more initial examples,  VI can enhance this diversity more effectively and consequently learn much richer image patterns, thereby providing more value. It is worth noting that our VI strategy is also compatible with other state-of-the-art particle detection frameworks. For example, when applying VI on DeepETPicker, we improve the F1 score from  0.276 to 0.351 on the N=10 setting.

\section{Conclusion}

In this work, we proposed the SaSi, a Self-augmented and Self-Interpreted framework for few-shot particle detection in cryo-ET data, incorporating novel strategies such as volume infilling and consistency guidance to address the challenges posed by limited training data. We demonstrated that our approach could enhance particle localization while maintaining robustness across different few-shot setups by utilizing a combination of supervised and self-supervised learning strategies. The integration of cc3d post-processing further improves the reliability of the detection process by eliminating the need for hyperparameter tuning.
We show that the core components of SaSi are compatible with other state-of-the-art frameworks.
The results across various configurations confirm the effectiveness of SaSi, including real-world datasets, and establish the benchmark for few-shot particle picking of Cryo-ET particles.
\section{Acknowledgments}
We thank Dr. Hideo Saito for his advice. This work was supported in part by U.S. NIH grants R01GM134020.


%
%
%





\bibliographystyle{plos2015}
\bibliography{cite}

\end{document}